\documentclass[conference]{IEEEtran}
\IEEEoverridecommandlockouts
\usepackage{cite}
\usepackage{amsmath,amssymb,amsfonts}
\usepackage{algorithmic}
\usepackage{graphicx}
\usepackage{textcomp}
\usepackage{xcolor}
\usepackage{enumitem}
\usepackage{float}
\usepackage[ruled,vlined]{algorithm2e}
\usepackage{amsmath}
\usepackage{url}

\def\BibTeX{{\rm B\kern-.05em{\sc i\kern-.025em b}\kern-.08em
    T\kern-.1667em\lower.7ex\hbox{E}\kern-.125emX}}
\begin{document}

\title{Object detection in adverse weather conditions for autonomous vehicles using Instruct Pix2Pix.}


\author{
\IEEEauthorblockN{
Unai Gurbindo\textsuperscript{$\dagger,\ddagger$},
Axel Brando\textsuperscript{$\dagger$},
Jaume Abella\textsuperscript{$\dagger$}, and
Caroline König\textsuperscript{$\ddagger$}
}
\IEEEauthorblockA{
\textsuperscript{$\dagger$}High-Performance Embedded Systems (HPES) Lab, Barcelona Supercomputing Center, Barcelona, Spain\\
\textsuperscript{$\ddagger$}Computer Science Dept., Univ. Politècnica de Catalunya, 
IDEAI-UPC- Research Center, Barcelona, Spain\\
\{unai.gurbindo, axel.brando, jaume.abella\}@bsc.es, ckonig@cs.upc.edu
}
}

\maketitle

\begin{abstract}
Enhancing the robustness of object detection systems under adverse weather conditions is crucial for the advancement of autonomous driving technology. This study presents a novel approach leveraging the diffusion model Instruct Pix2Pix to develop prompting methodologies that generate realistic datasets with weather-based augmentations aiming to mitigate the impact of adverse weather on the perception capabilities of state-of-the-art object detection models, including Faster R-CNN and YOLOv10. Experiments were conducted in two environments, in the CARLA simulator where an initial evaluation of the proposed data augmentation was provided, and then on the real-world image data sets BDD100K and ACDC demonstrating the effectiveness of the approach in real environments.

The key contributions of this work are twofold: (1) identifying and quantifying the performance gap in object detection models under challenging weather conditions, and (2) demonstrating how tailored data augmentation strategies can significantly enhance the robustness of these models. This research establishes a solid foundation for improving the reliability of perception systems in demanding environmental scenarios, and provides a pathway for future advancements in autonomous driving.
\end{abstract}

\begin{IEEEkeywords}
 Adverse Weather Conditions, Autonomous Driving, Data Augmentation, Denoising Models, Diffusion Models, Object Detection, Robustness, Instruct Pix2Pix.  
\end{IEEEkeywords}

\section{Introduction}

Autonomous driving is one of the most significant technological advancements of the last decade, with the potential to radically transform transportation and urban mobility.

This progress has been driven by rapid developments in artificial intelligence, machine learning, and computer vision, aiming to reduce accidents, enhance traffic efficiency, and provide mobility access to individuals with disabilities or those without access to traditional vehicles. 

Despite these promising advancements, autonomous driving systems face several critical challenges that hinder their implementation and widespread adoption \cite{intro_driving_1}.

\begin{figure}[htbp]
\centering
\includegraphics[width=.4677\textwidth]{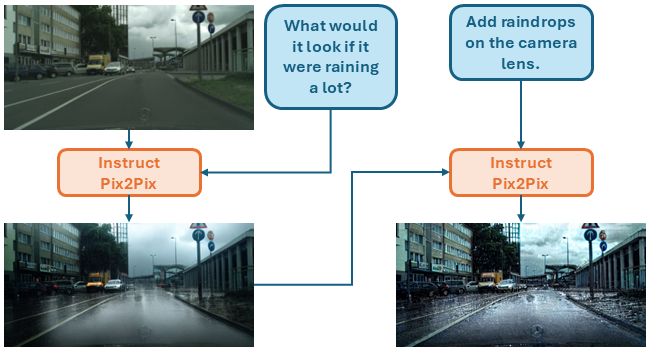}
\caption{Example of a Prompting Strategy to Simulate Rain Effect on an Image Using Two Sequentially Applied Prompts.}
\label{fig:fig1}
\end{figure}

Among these challenges, adverse weather conditions stand out as a critical area for improvement due to their impact on sensor reliability and overall safety. This article focuses on enhancing the performance of artificial vision systems in Intelligent Vehicles (IV), specifically those responsible for detecting objects on the road (for example, vehicles, traffic lights, and road signs), to ensure robust functionality under adverse weather conditions.

Currently, object detection systems can be categorized into two main types: one-stage models and two-stage models (also named region-based models). The most well-known examples are Faster Region-Based Convolutional Neural Networks (Faster-RCNN) \cite{faster_rcnn} and the recent You Only Look Once (YOLO) version 10 \cite{yolov10}. However, these models face challenges due to the unpredictable and complex nature of everyday driving scenarios, especially those caused by adverse weather conditions. One reason for this, as highlighted in \cite{intro_driving_1}, is the difficulty in collecting datasets that accurately represent these conditions.



To address these challenges, previous studies have proposed strategies such as data augmentation to improve dataset diversity. In this article, we present a data augmentation technique based on a sequential prompting strategy using diffusion models, particularly the Instruct Pix2Pix model \cite{instructpix2pix}. We propose a novel augmentation method that generates realistic weather conditions while preserving spatial and semantic consistency, which is essential for building effective object models. In addition, we provide a comprehensive quantitative evaluation showing that our approach improves the robustness of object detection systems under adverse weather scenarios.

\section{Background}

\subsection{Development Process of Automotive Systems}

The development process of safety-relevant automotive systems (e.g., braking, steering, etc.) must adhere to specific regulations, such as ISO 26262~\cite{ISO26262} and ISO 21448~\cite{ISO21448}. Whenever those systems include AI components with some impact on safety requirements, as it is the case for Advanced Driver Assistance Systems (ADAS) and autonomous vehicles, then specific extensions of those regulations must be accounted for, such as the development process provided in ISO/PAS 8800~\cite{ISO8800}, whose main characteristics are publicly detailed by Fernandez et al.~\cite{CARS24}. In particular, one key step when deploying neural networks in safety-critical automotive systems is ensuring completeness, representativeness, balance and volume for the data used to train and validate the neural network. This applies to each of the driving scenarios that must be accounted for, with those scenarios being identified following specific methodologies during the development process of the system, as prescribed by the relevant standards.

Unfortunately, collecting analogous data in different weather conditions is difficult, and for particularly scarce data, such as for instance videos prior to severe crashes, virtually impossible. In general, the vast majority of data available is collected in favorable weather conditions, which challenges achieving the desired completeness, representativeness, balance and volume mainly due to the lack of sufficient data under adverse weather conditions. Therefore, solutions for data augmentation are extremely convenient to tackle this challenge.

\subsection{Object Detection Models}

Object detection models are generally divided into two main categories: one-stage models (which predict the bounding box and classify it simultaneously) and two-stage models, also known as region-based models (which first propose object regions and then classify them). In this context, YOLO \cite{yolov10} and Faster R-CNN \cite{faster_rcnn} have emerged as key benchmarks due to their efficient approach and innovative features.

\textbf{YOLOv10} model is the result of the evolution of several previous versions \cite{yolo_survey}. The main idea behind this model can be traced back to its first version, YOLOv1, where the concept was to divide an image into a fixed-size grid, with each cell responsible for predicting the bounding box containing the object and its category. This simplified design combines detection and classification into a single step, offering significantly higher speed compared to two-stage methods.

YOLO models have rapidly evolved, reaching version 10, YOLOv10, which represents a significant advancement within the series. It stands out for eliminating the need for traditional post-processing with Non-Maximum Suppression (NMS). This is achieved through an innovative one-to-one prediction mechanism, where the inference block is directly supervised by another block responsible for generating one-to-many predictions during training. This supervised approach improves accuracy and efficiency by reducing redundancies in detections. YOLOv10’s structure remains consistent with the general YOLO design, consisting of three main components: the backbone, responsible for feature extraction; the neck, which merges these features; and the head, which generates the final detections.

\textbf{Faster R-CNN} is a two-stage object detection model widely recognized for its accuracy and flexibility. It introduces a Region Proposal Network (RPN) that efficiently generates proposals for relevant regions, significantly reducing computational cost compared to its predecessors, such as Fast R-CNN \cite{faster_rcnn}. In its first stage, Faster R-CNN identifies regions of interest, and in the second stage, it refines these regions to classify objects and adjust bounding boxes. Although highly accurate, its two-stage nature makes it less efficient in terms of speed compared to one-stage models like YOLO, limiting its applicability in real-time scenarios.


Both approaches, YOLO and Faster R-CNN, offer distinct solutions addressing different needs within the field of object detection.

\subsection{Diffusion Models}

Diffusion models have emerged as a powerful family of generative models for creating high-quality images, audio, and other data modalities \cite{yang2024diffusionmodelscomprehensivesurvey}. The objective of these models is to gradually add noise to the data during the forward process and learn to reverse this process step-by-step in the backward process to generate new samples from pure noise. The foundational equations are as follows \cite{dm_graph}:

1. \textbf{Forward Process (noising):}
\begin{equation}  
\begin{aligned}
q(\mathbf{x}_{1:T} | \mathbf{x}_0) &:= \prod_{t=1}^T q(\mathbf{x}_t | \mathbf{x}_{t-1}), \\
q(\mathbf{x}_t | \mathbf{x}_{t-1}) &:= \mathcal{N}(\mathbf{x}_t; \sqrt{1 - \beta_t} \, \mathbf{x}_{t-1}, \beta_t \mathbf{I}).
\end{aligned}
\label{eq:q}
\end{equation}
where $\beta_t$ controls the noise schedule.

2. \textbf{Reverse Process (denoising):}
\begin{equation}  
\begin{aligned}
p_\theta(\mathbf{x}_{0:T}) &:= p(\mathbf{x}_T) \prod_{t=1}^T p_\theta(\mathbf{x}_{t-1}|\mathbf{x}_t), \\ 
p_\theta(\mathbf{x}_{t-1}|\mathbf{x}_t) &:= \mathcal{N}(\mathbf{x}_{t-1}; \boldsymbol{\mu}_\theta(\mathbf{x}_t, t), \boldsymbol{\Sigma}_\theta(\mathbf{x}_t, t)).
\end{aligned}
\label{eq:p0}
\end{equation}
where $\mu_\theta$ and $\Sigma_\theta$ are learned parameters.


The combination of these processes allows the model to reverse the noise-adding process and generate realistic samples.

This noise removal process can be conditioned to guide the generation toward specific attributes or signals. For instance, one common approach is classifier-free guidance, where the model is trained with and without conditioning information. During inference, a guidance scale can be applied to adjust the influence of the conditioning, enabling tasks such as text-guided image generation or domain-specific data synthesis.


Building on this, the \textbf{Instruct Pix2Pix} model \cite{instructpix2pix} expands the capabilities of diffusion models by introducing image and text conditioning. Based on the Stable Diffusion \cite{stable_diffusion} framework---a latent diffusion model designed for high-quality image generation from text prompts---Instruct Pix2Pix allows localized edits to images using textual instructions.

By leveraging a pre-trained latent space and incorporating both an input image and textual guidance, Instruct Pix2Pix demonstrates the flexibility of diffusion models. This approach excels in tasks requiring both creative generation and precise editing, making it a valuable tool for real-world applications.

\section{Materials}

The materials used in this study consist of images capturing road scenarios in which traffic signs, traffic lights, vehicles, and pedestrians are present, along with their respective bounding boxes. These images are obtained under various weather conditions to analyze the central focus of this study: the sensitivity of detection systems to differing meteorological conditions.  

Two types of experiments are conducted: one with simulated images, providing an initial assessment of data augmentation, and another with real images, demonstrating the feasibility of this technique in real-world environments. The following sections detail the origin of the datasets used in the project.

\subsection{Simulated Data}

The simulated data in this study came from images rendered in the open source simulator CARLA \cite{carla_simulator}, a simulator that provides detailed control of dynamic driving scenarios and includes the ability to adjust factors such as weather conditions. 

Object bounding box annotations were extracted using information integrated in CARLA from objects in its environment. A filtering step was necessary to address the ``ghost boxes” caused by occlusions by comparing the RGB and segmentation images pixel by pixel \cite{carla_box}, thus obtaining a set of simulated images with their corresponding annotations.

\subsection{Real-World Data}

The real data for this project come from two real-image datasets. The first one, BDD100K \cite{BDD100K}, contains 100,000 labeled images from diverse driving scenarios with annotations for multiple classes of objects, weather conditions, and time of day. While this broad coverage is valuable for evaluating detection in various environments, it offers relatively few images for certain weather conditions such as fog, which limits the representation of some specific weather scenarios.

The second dataset, ACDC \cite{acdc_data}, was created specifically to evaluate systems under adverse weather conditions, such as rain, fog, snow, and nighttime scenes. In this dataset, traffic lights and traffic signs are not annotated. To overcome this limitation, the semantic segmentation masks provided by the dataset were used to generate bounding boxes for these classes using the \texttt{skimage.measure.regionprops} function \cite{scikit-image}.

The use of these two datasets provides a solid basis for studying detection performance under various weather conditions in real-world settings.

\section{Methodology}

This section describes the methodological framework employed to improve the robustness of object detection in adverse weather conditions using the Instruct Pix2Pix diffusion model.  

The methodology is based on the development of a data augmentation strategy, which is applied to two data frames: the simulated data frame and the real-world data frame. Each framework leverages augmentation strategies to generate augmentation based on weather conditions, thereby increasing the diversity and realism of the training datasets.  

The following subsections provide an overview of the processes, prompts, and hyperparameters used in each framework.

\subsection{Data Augmentation Pipeline}

The data augmentation process in both frameworks is designed to systematically enhance the diversity of the dataset by simulating adverse weather conditions. The pipeline, as outlined in Algorithm \ref{alg:pix2pix}, leverages the capabilities of the Instruct Pix2Pix model to modify clear-weather images while ensuring that the augmented dataset maintains relevance and realism.

\begin{algorithm}[t]
\small
\caption{Data Augmentation Pipeline Using Instruct Pix2Pix}
\label{alg:pix2pix}
\KwIn{%
  A dataset $\mathcal{D}$ of images without adverse weather conditions}
\KwOut{%
  Augmented dataset $\mathcal{D'}$ containing images with diverse weather-induced modifications}
\BlankLine
\textbf{Step 1: Prompt Application} \\
Apply the Instruct Pix2Pix model to every image in $\mathcal{S}$ using predefined prompts and hyperparameters for each target weather condition. In some weather conditions, different prompts are required to be added sequentially (Example in Figure \ref{fig:fig1}). \\
\BlankLine
\textbf{Step 2: Augmented Image Generation} \\
Generate augmented images $\mathcal{D'}$ containing the specified weather-induced modifications. \\
\BlankLine
\textbf{Step 3: Filtering Step} \\
Perform a manual filtering step on $\mathcal{D'}$ to remove:
\begin{enumerate}
  \item Augmented images with missing objects due to model hallucinations.
  \item Augmented images exhibiting extreme or unrealistic perturbations.
\end{enumerate}
Integrate remaining filtered images into the original dataset, forming the final dataset $\mathcal{D'} = \mathcal{D} \cup \mathcal{D'}_{\text{filtered}}$, where both clear-weather images and those with adverse weather conditions are available. \\
\Return $\mathcal{D'}$
\end{algorithm}

\subsection{Implementation Details}

Two key hyperparameters are used in this work: the \emph{guidance scale} (defined as \(s_T\) in \cite{instructpix2pix}) and the number of \emph{inference steps}. The guidance scale used in this study controls how strictly the model adheres to the prompt description; higher values result in stronger stylistic changes but may reduce fidelity to the original content. The number of inference steps determines how many iterations the diffusion process runs, with higher values improving image quality and detail at the cost of computational efficiency. Balancing these parameters is crucial to achieving visually coherent and realistic transformations.

Specific prompts and hyperparameters used for each weather condition and data framework are summarized below.





\paragraph{Simulated Framework Prompting Details}
\begin{itemize}[leftmargin=*,labelsep=5pt]
    \item \textbf{Rain:}
        \begin{enumerate}[label=\arabic*., leftmargin=2em]
            \item \textit{What would it look if it were raining a lot?}\\
                  \textbf{Guidance Scale}: 1.45,\quad
                  \textbf{Inference Steps}: 100.
            \item \textit{Add raindrops on the camera lens.}\\
                  \textbf{Guidance Scale}: 1.65,\quad
                  \textbf{Inference Steps}: 100.
        \end{enumerate}

    \item \textbf{Fog:}
        \begin{enumerate}[label=\arabic*., leftmargin=2em]
            \item \textit{Add dense fog to the image.}\\
                  \textbf{Guidance Scale}: 1.9,\quad
                  \textbf{Inference Steps}: 100.
        \end{enumerate}

    \item \textbf{Night:}
        \begin{enumerate}[label=\arabic*., leftmargin=2em]
            \item \textit{What would it look like at night?}\\
                  \textbf{Guidance Scale}: 1.5,\quad
                  \textbf{Inference Steps}: 100.
            \item \textit{Add a lot of darkness.}\\
                  \textbf{Guidance Scale}: 1.75,\quad
                  \textbf{Inference Steps}: 100.
        \end{enumerate}
\end{itemize}

\vspace{1em}
\paragraph{Real-World Data Prompting Details}
\begin{itemize}[leftmargin=*,labelsep=5pt]
    \item \textbf{Rain:}
        \begin{enumerate}[label=\arabic*., leftmargin=2em]
            \item \textit{What would it look if it were raining a lot?}\\
                  \textbf{Guidance Scale}: 1.35,\quad
                  \textbf{Inference Steps}: 250.
            \item \textit{Add raindrops on the camera lens.}\\
                  \textbf{Guidance Scale}: 2,\quad
                  \textbf{Inference Steps}: 200.
        \end{enumerate}

    \item \textbf{Fog:}
        \begin{enumerate}[label=\arabic*., leftmargin=2em]
            \item \textit{Add dense fog to the image.}\\
                  \textbf{Guidance Scale}: 1.9,\quad
                  \textbf{Inference Steps}: 100.
        \end{enumerate}

    \item \textbf{Night:}
        \begin{enumerate}[label=\arabic*., leftmargin=2em]
            \item \textit{What would it look like at night?}\\
                  \textbf{Guidance Scale}: 1.5,\quad
                  \textbf{Inference Steps}: 100.
            \item \textit{Add a lot of darkness.}\\
                  \textbf{Guidance Scale}: 1.75,\quad
                  \textbf{Inference Steps}: 100.
        \end{enumerate}

    \item \textbf{Snow:}
        \begin{enumerate}[label=\arabic*., leftmargin=2em]
            \item \textit{What would it look like were snowing?}\\
                  \textbf{Guidance Scale}: 1.25,\quad
                  \textbf{Inference Steps}: 150.
            \item \textit{Add snowflakes falling from the sky.}\\
                  \textbf{Guidance Scale}: 1.5,\quad
                  \textbf{Inference Steps}: 125.
        \end{enumerate}
\end{itemize}

The previous prompt formulations are inspired by examples from the original Instruct Pix2Pix paper, which demonstrated strong results with inputs such as \textit{What would it look like if it were snowing?} and \textit{Add fireworks to the sky} (Figure 1 in \cite{instructpix2pix}). Moreover, applying prompts sequentially—rather than simultaneously—has been shown to improve prompting effect, as illustrated in Figure 11 of \cite{instructpix2pix}. Also, to preserve semantic integrity, the guidance scale has been deliberately kept low to reduce hallucinations. These design choices may make
our augmentation approach suitable for other contexts as well.


\begin{figure}[t]
\centering
\includegraphics[width=.465\textwidth]{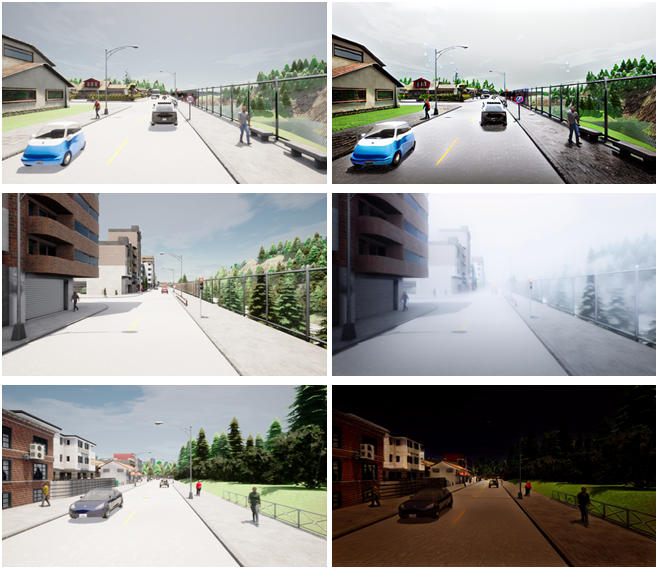}
\caption{Example of data augmentation in a simulated data framework to generate adverse weather conditions such as rain, fog, and nighttime. The left images represent the inputs of the process, while the right images show the augmented outputs with the same scenes under adverse weather conditions.}
\label{fig:data_aug_sim}
\end{figure}

\begin{figure}[t]
\centering
\includegraphics[width=.469\textwidth]{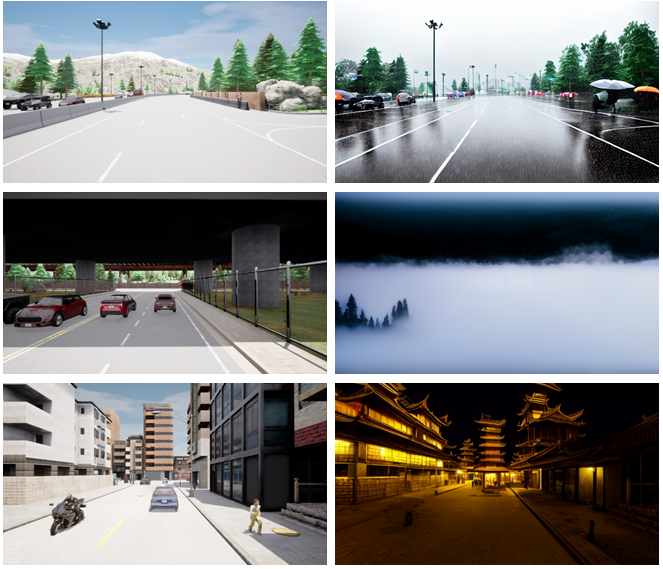}
\caption{Example of hallucinations generated and corrected after applying data augmentation in the simulated framework for weather conditions of rain, fog, and nighttime. The images on the left represent the input of the process, while the images on the right show the augmented images, some of which exhibit hallucinations during the inference process.}
\label{fig:hallucinations_simu}
\end{figure}

\begin{figure}[t]
\centering
\includegraphics[width=.465\textwidth]{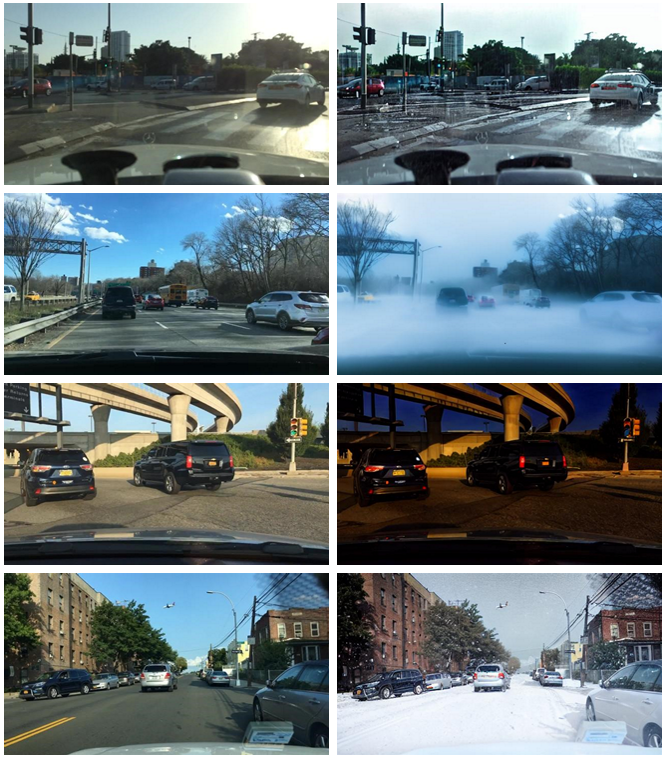}
\caption{Example of data augmentation in a real-world data framework to generate adverse weather conditions such as rain, fog, nighttime and snow. The left images represent the inputs of the process, while the right images show the augmented outputs with the same scenes under adverse weather conditions.}
\label{fig:da}
\end{figure}

\begin{figure}[t]
\centering
\includegraphics[width=.455\textwidth]{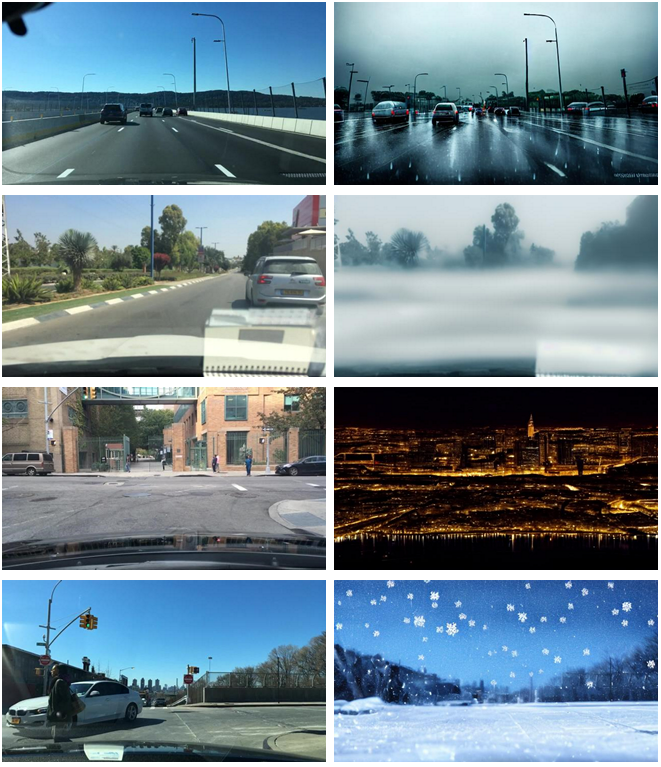}
\caption{Example of hallucinations generated and corrected after applying data augmentation in the real-world data framework for weather conditions of rain, fog, nighttime and snow. The images on the left represent the input of the process, while the images on the right show the augmented images, some of which exhibit hallucinations during the inference process.}
\label{fig:hal}
\end{figure}

\section{Experiments}

\subsection{Datasets}

To evaluate the prompting strategy for generating data augmentation, two datasets are created for each data framework. The first dataset consists of images without adverse climatic conditions (the \textit{Basic} dataset), while the second dataset includes a mix of images referred to as the \textit{Augmented} dataset. In this second dataset, a percentage of the images are taken from the original dataset, and the remaining images are artificially generated using the prompting strategy.

\textbf{Simulated Datasets:} In this case, a total of 5,064 road scenarios are generated in the CARLA simulator. Of these, 254 are reserved to form the test set used to evaluate the experiments in this framework. These scenarios are rendered not only in daytime situations without adverse climatic conditions (referred to as ``default" in this work) but also under rain, fog, and night, the three adverse climatic conditions studied in this framework. The objective is to create a test set for the experiments within this framework to evaluate the performance of the models under each meteorological condition. Thus, the test set consists of 1,016 images, with 254 images for each climatic condition.  

As for the remaining, these are rendered exclusively under default conditions and are divided into 70\% (3,367 scenarios) for training and 30\% (1,443 scenarios) for validation of the object detection models. Therefore, the training dataset for each approach consists of a total of 4,810 images. One of the datasets in this framework, referred to as the \textit{Basic} dataset, includes only default-condition images. The other dataset, incorporating data augmentation techniques from this study, consists of 40\% default images from the simulator (1,348 images for training and 579 for validation), while the remaining 60\% are artificially generated to represent adverse weather conditions. Specifically, 20\% of the dataset is allocated to each weather condition (673 images per condition for training and 288 for validation).

\textbf{Real-World Datasets:} In this case, 7,083 images are selected from the BDD100K dataset, all captured during daytime with clear weather. These images are split into 70\% for training and 30\% for validation of the object detection models. The dataset constructed with augmented images in this framework follows a balanced 50\%-50\% distribution, where half of the images come from the original dataset, and the other half are artificially generated under adverse weather conditions. Unlike the simulated framework, this dataset includes information about snow, a climatic condition that cannot be rendered in the CARLA simulator. Consequently, the \textit{Augmented} dataset consists of 2,500 default images for training and 1,043 for validation, while for each adverse climatic condition—rain, fog, and snow—there are 625 training images and 260 validation images.

The test set for real-world datasets used to evaluate the experiments is constructed using the ACDC dataset, selecting 250 images for each of the four adverse weather conditions: rain, fog, night, and snow. Since this dataset does not include images under default conditions, an additional 250 images are taken from the BDD100K dataset. As a result, the test set for this framework consists of a total of 1,250 images, with 250 images representing each weather condition.

\subsection{Training Details}

\textbf{Faster-RCNN Specific:} The selected Faster R-CNN model employs a ResNet-50 backbone with pre-trained weights from the COCO dataset and no architectural modifications. It trains using a batch size of 8, a learning rate of \(1 \times 10^{-4}\), and for 10 epochs, with the batch size maximized based on the available hardware. A low learning rate is chosen to effectively adapt the pre-trained model to the specific images of the problem.

\textbf{YOLOv10 Specific:} The selected YOLOv10 models, Nano and Medium, use pre-trained weights from the COCO dataset, similar to the Faster R-CNN model. Employing two different model sizes enables the analysis of data augmentation effects across varying scales and the evaluation of its impact on performance. The training parameters are set with a batch size of 8, a learning rate of \(1 \times 10^{-3}\), and 100 epochs. This configuration enables a comprehensive evaluation of the YOLOv10 models under the specified training conditions.

\subsection{Experimental Settings}

For each framework, the quality of data augmentation achieved through the \textit{prompting} strategy and the subsequent removal of ``hallucinations'' (irrelevant or erroneous images) is analyzed.

The selected object detection models are then trained using two datasets: a basic dataset, which does not contain any images representing adverse weather, and another that incorporates the data augmentation strategy.

Next, each model is evaluated separately for each weather condition by means of a resampling method that relies on the test set defined for each framework. This resampling procedure draws 1,000 bootstrap samples of the same size as the original dataset for each weather condition, using sampling with replacement. In every iteration, inference is performed on the selected images, and the mean Average Precision (\(\text{mAP}\)) metric is obtained, where \(\text{mAP}\) is derived from the Average Precision (\(\text{AP}\)) for each object class at a threshold of \(0.5\). After computing \(\text{mAP}\) values for all 1,000 resamples, the mean and standard deviation of these values are calculated to evaluate overall model performance.

For example, in the simulated framework, each weather condition is represented by 250 images. To evaluate model performance under one of these conditions, 1,000 bootstrap samples of 250 images each are drawn with replacement from the images of the condition under analysis. Inference is then performed on each sample, and the \(\text{mAP}\) and \(\text{AP}\) values are recorded across the 1,000 iterations. The mean and standard deviation of both \(\text{mAP}\) and \(\text{AP}\) are then computed from these iterative results to summarize performance.

This methodology is selected because models are not entirely deterministic, and the number of object types can differ across weather conditions in the test sets. These variations arise in both simulated and real-world data frameworks, which motivates the adoption of a resampling strategy to produce stable performance estimates.

\subsection{Results and Discussion}

\textbf{Simulated Data Experiments:} Considering the number of images available in this framework, data augmentation is applied to the training and validation images of the \textit{Basic} dataset, resulting in a total of 4,810 images for the three climatic conditions analyzed in this framework (examples of this data augmentation can be seen in Figure \ref{fig:data_aug_sim}). However, as mentioned earlier, it is necessary to filter the images due to hallucinations sometimes caused by the Instruct Pix2Pix model (examples of hallucinations in Figure \ref{fig:hallucinations_simu}). After the filtering process, the final dataset includes 4,270 fog images, 4,484 rain images, and 4,780 night images. 

Once the \textit{Basic} and \textit{Augmented} datasets for the simulated images are created, the respective models (Faster R-CNN, YOLO SIZE M/N) are trained on these datasets. Tables~\ref{tab:results_faster_rcnn_map50_sim}--\ref{tab:results_yolo_size_n_map50_sim} present the models' performance according to the mAP50 metric for each weather condition.

\begin{table}[ht!]
\centering
\scriptsize
\caption{Results of Faster RCNN evaluated with mAP50 (mean ± std) in the Simulated Framework.}
\begin{tabular}{|c|c|c|}
\hline
\multicolumn{1}{|c|}{\textbf{Weather}} & \multicolumn{2}{c|}{\textbf{Approaches}} \\ \cline{2-3} 
\textbf{Condition} & Augmented & Basic \\ \hline
Default & 0.838 ± 0.01 & \textbf{0.881 ± 0.008} \\
Fog & \textbf{0.781 ± 0.013} & 0.635 ± 0.015 \\
Night & \textbf{0.717 ± 0.014} & 0.556 ± 0.017 \\
Rain & \textbf{0.688 ± 0.015} & 0.68 ± 0.015 \\
\hline
\end{tabular}
\label{tab:results_faster_rcnn_map50_sim}
\end{table}

\begin{table}[ht!]
\caption{Results of YOLO Size M evaluated with mAP50 (mean ± std) in the Simulated Framework.}
\centering
\scriptsize
\begin{tabular}{|c|c|c|}
\hline
\multicolumn{1}{|c|}{\textbf{Weather}} & \multicolumn{2}{c|}{\textbf{Approaches}} \\ \cline{2-3} 
\textbf{Condition} & Augmented & Basic \\ \hline
Default & \textbf{0.55 ± 0.015} & 0.536 ± 0.015 \\
Fog & \textbf{0.493 ± 0.016} & 0.395 ± 0.015 \\
Night &\textbf{ 0.457 ± 0.016} & 0.336 ± 0.013 \\
Rain & \textbf{0.435 ± 0.016} & 0.38 ± 0.015 \\
\hline
\end{tabular}
\label{tab:results_yolo_size_m_map50_sim}
\end{table}

\begin{table}[ht!]
\centering
\caption{Results of YOLO Size N evaluated with mAP50 (mean ± std) in the Simulated Framework.}
\scriptsize
\begin{tabular}{|c|c|c|}
\hline
\multicolumn{1}{|c|}{\textbf{Weather}} & \multicolumn{2}{c|}{\textbf{Approaches}} \\ \cline{2-3} 
 \textbf{Condition} & Augmented & Basic \\ \hline
Default & 0.408 ± 0.012 & \textbf{0.422 ± 0.014} \\
Fog & \textbf{0.391 ± 0.013} & 0.302 ± 0.014 \\
Night &\textbf{ 0.337 ± 0.013} & 0.266 ± 0.012 \\
Rain & \textbf{0.312 ± 0.013} & 0.236 ± 0.013 \\
\hline
\end{tabular}
\label{tab:results_yolo_size_n_map50_sim}
\end{table}
These results show that for the default condition, the Faster R-CNN model and the YOLO Nano model achieve better results with the \textit{Basic} approach, which is expected because the training process focuses exclusively on this condition.

However, for adverse weather conditions, the \textit{Augmented} approach yields better results in all models, thus improving robustness. The difference in performance between default and adverse weather conditions is clearly smaller than with the \textit{Basic} approach. 


\begin{table}[h!]
\centering
\caption{Results of Faster RCNN evaluated with AP50 (mean ± std) in the Simulated Framework by weather conditions.}
\scriptsize
\setlength{\tabcolsep}{0.65mm}
\begin{tabular}{|c|c|c|c|c|c|}
\hline
\textbf{Weather} & \textbf{Approaches} & \multicolumn{4}{c|}{\textbf{Object Class}} \\ \cline{3-6} 
\textbf{Condition} &  & Walker & Vehicle & Traffic  & Traffic  \\
 &  &  &  &  Signs &  Lights \\ 
\hline
Default & Augmented & 0.755 ± 0.015 & 0.910 ± 0.009 & 0.816 ± 0.031 & 0.874 ± 0.016 \\
 & Basic & \textbf{0.780 ± 0.014} & \textbf{0.923 ± 0.008} & \textbf{0.889 ± 0.025} & \textbf{0.933 ± 0.010} \\
\hline
Fog & Augmented & \textbf{0.673 ± 0.019} & \textbf{0.850 ± 0.011} & \textbf{0.749 ± 0.037} & \textbf{0.852 ± 0.028} \\
 & Basic & 0.469 ± 0.021 & 0.694 ± 0.013 & 0.606 ± 0.043 & 0.769 ± 0.031 \\
\hline
Night & Augmented & \textbf{0.627 ± 0.018} & \textbf{0.872 ± 0.010} & \textbf{0.673 ± 0.042} & \textbf{0.695 ± 0.026} \\
 & Basic & 0.488 ± 0.021 & 0.669 ± 0.017 & 0.535 ± 0.048 & 0.532 ± 0.032 \\
\hline
Rain & Augmented & \textbf{0.507 ± 0.021} & \textbf{0.862 ± 0.010} & 0.678 ± 0.044 & 0.706 ± 0.032 \\
 & Basic & 0.464 ± 0.023 & 0.837 ± 0.010 & \textbf{0.693 ± 0.042} & \textbf{0.727 ± 0.031} \\
\hline
\end{tabular}
\label{tab:results_faster_rcnn_per_class_ap50_95_per_weather_sim}
\end{table}

\begin{table}[h!]
\centering 
\caption{Results of YOLO Size M evaluated with AP50 (mean ± std) in the Simulated Framework by weather conditions.}
\scriptsize
\setlength{\tabcolsep}{0.65mm}
\centering
\begin{tabular}{|c|c|c|c|c|c|}
\hline
\textbf{Weather} & \textbf{Approaches} & \multicolumn{4}{c|}{\textbf{Object Class}} \\ \cline{3-6} 
\textbf{Condition} &  & Walker & Vehicle & Traffic  & Traffic  \\
 &  &  &  &  Signs &  Lights \\ \hline
Default & Augmented & \textbf{0.497 ± 0.022} & \textbf{0.833 ± 0.012} & \textbf{0.433 ± 0.043} & 0.436 ± 0.032 \\
 & Basic & 0.487 ± 0.021 & 0.825 ± 0.012 & 0.389 ± 0.043 & \textbf{0.443 ± 0.030} \\
\hline
Fog & Augmented & \textbf{0.416 ± 0.021} & \textbf{0.732 ± 0.014} & 0.417 ± 0.046 & \textbf{0.408 ± 0.034} \\
 & Basic & 0.316 ± 0.018 & 0.509 ± 0.015 & \textbf{0.431 ± 0.045} & 0.324 ± 0.034 \\
\hline
Night & Augmented & \textbf{0.370 ± 0.021} & \textbf{0.746 ± 0.013} & \textbf{0.335 ± 0.045} & \textbf{0.378 ± 0.032} \\
 & Basic & 0.245 ± 0.017 & 0.571 ± 0.016 & 0.254 ± 0.038 & 0.276 ± 0.023 \\
\hline
Rain & Augmented & \textbf{0.285 ± 0.021} & \textbf{0.724 ± 0.013} & \textbf{0.366 ± 0.045} & \textbf{0.366 ± 0.036} \\
 & Basic & 0.191 ± 0.018 & 0.658 ± 0.013 & 0.358 ± 0.045 & 0.313 ± 0.032 \\
\hline
\end{tabular}
\label{tab:results_yolo_size_m_per_class_ap50_95_per_weather_sim}
\end{table}


\begin{table}[h!]
\centering
\caption{Results of YOLO Size N evaluated with AP50 (mean ± std) in the Simulated Framework by weather conditions.}
\scriptsize
\setlength{\tabcolsep}{0.65mm}
\begin{tabular}{|c|c|c|c|c|c|}
\hline
\textbf{Weather} & \textbf{Approaches} & \multicolumn{4}{c|}{\textbf{Object Class}} \\ \cline{3-6} 
\textbf{Condition} &  & Walker & Vehicle & Traffic  & Traffic  \\
 &  &  &  &  Signs &  Lights \\ 
\hline
Default & Augmented & 0.353 ± 0.023 & 0.720 ± 0.012 & \textbf{0.294 ± 0.034} & 0.264 ± 0.026 \\
        & Basic     & \textbf{0.369 ± 0.021} & \textbf{0.725 ± 0.011} & 0.272 ± 0.035 & \textbf{0.322 ± 0.033} \\
\hline
Fog     & Augmented & \textbf{0.306 ± 0.020} & \textbf{0.612 ± 0.014} & \textbf{0.328 ± 0.035} & \textbf{0.319 ± 0.029} \\
        & Basic     & 0.241 ± 0.019 & 0.460 ± 0.014 & 0.283 ± 0.040 & 0.222 ± 0.028 \\
\hline
Night   & Augmented & \textbf{0.274 ± 0.019} & \textbf{0.627 ± 0.014} & \textbf{0.244 ± 0.036} & \textbf{0.204 ± 0.024} \\
        & Basic     & 0.210 ± 0.017 & 0.468 ± 0.017 & 0.212 ± 0.035 & 0.173 ± 0.022 \\
\hline
Rain    & Augmented & \textbf{0.195 ± 0.018} & \textbf{0.593 ± 0.013} & \textbf{0.263 ± 0.038} & \textbf{0.198 ± 0.025} \\
        & Basic     & 0.113 ± 0.014 & 0.446 ± 0.015 & 0.198 ± 0.036 & 0.187 ± 0.028 \\
\hline
\end{tabular}
\label{tab:results_yolo_size_n_per_class_ap50_95_per_weather_sim}
\end{table}

Tables~\ref{tab:results_faster_rcnn_per_class_ap50_95_per_weather_sim}--\ref{tab:results_yolo_size_n_per_class_ap50_95_per_weather_sim} present the performance of the models for specific object types and weather conditions. These results align with the previous observation, showing that the data augmentation approach typically outperforms the \textit{Basic} approach in adverse climatic conditions. An improvement in vehicle detection is particularly noticeable, where the gap between the \textit{Augmented} and \textit{Basic} approaches is marked. For example, in the YOLO model of size~M, the performance difference in detecting vehicles between the default and foggy conditions is around 0.3 for the \textit{Basic} approach, while it is only 0.1 with the \textit{Augmented} approach. This outcome further highlights the potential of the data augmentation strategy to enhance the robustness of object detection models.

\textbf{Real-World Data Experiments Results:} A total of 7,083 images are available for training and validation. Due to computational constraints, data augmentation is carried out on 4,247 images across four climatic conditions (examples of some augmentations appear in Figure~\ref{fig:da}). After filtering out hallucinations (examples of hallucinations in Figure~\ref{fig:hal}), the final dataset contains 2,981 images for fog, 3,449 for rain, 4,137 for night, and 3,799 for snow, images with which the \textit{Augmented} dataset of this framework is created. Object detection models are then trained using this dataset and the basic one, and the resulting performance is evaluated.

Tables~\ref{tab:results_faster_rcnn_map50}--\ref{tab:results_yolo_size_n_map50} present the mAP metric for each model, revealing varied outcomes. Faster R-CNN generally shows an improvement when the data augmentation approach is applied, narrowing the performance gap between default and adverse weather conditions, though the improvement is not substantial. The YOLO models produce very similar results, indicating no clear enhancement in model robustness. These models also underperform compared to Faster R-CNN, suggesting a need for further investigation into their training process to achieve stable results, particularly under default conditions.

\begin{table}[ht!]
\centering
\scriptsize
\caption{Results of Faster RCNN evaluated with mAP50 (mean ± std) in the Real-World Data Framework.}
\begin{tabular}{|c|c|c|}
\hline
\multicolumn{1}{|c|}{\textbf{Weather Condition}} & \multicolumn{2}{c|}{\textbf{Approaches}} \\ \cline{2-3} 
 & Augmented & Basic \\ \hline
Default & \textbf{0.728 ± 0.014} & 0.635 ± 0.015 \\
Fog & \textbf{0.449 ± 0.017} & 0.429 ± 0.017 \\
Night & \textbf{0.413 ± 0.011} & 0.345 ± 0.011 \\
Rain & \textbf{0.375 ± 0.014} & 0.348 ± 0.012 \\
Snow & \textbf{0.435 ± 0.01} & 0.413 ± 0.01 \\
\hline
\end{tabular}
\label{tab:results_faster_rcnn_map50}
\end{table}

\begin{table}[ht!]
\centering
\scriptsize
\caption{Results of YOLO Size M evaluated with (mean ± std) in the Real-World Data Framework.}
\begin{tabular}{|c|c|c|}
\hline
\multicolumn{1}{|c|}{\textbf{Weather Condition}} & \multicolumn{2}{c|}{\textbf{Approaches}} \\ \cline{2-3} 
 & Augmented & Basic \\ \hline
Default & 0.372 ± 0.015 & \textbf{0.375 ± .0013} \\
Fog & \textbf{0.256 ± 0.014} & 0.255 ± 0.014 \\
Night & 0.169 ± 0.008 & \textbf{0.182 ± 0.008} \\
Rain & 0.182 ± 0.009 & \textbf{0.192 ± 0.011} \\
Snow & 0.215 ± 0.008 & \textbf{0.224 ± 0.009} \\
\hline
\end{tabular}
\label{tab:results_yolo_size_m_map50}
\end{table}

\begin{table}[ht!]
\centering
\scriptsize
\caption{Results of YOLO Size N evaluated with (mean ± std) in the Real-World Data Framework.}
\begin{tabular}{|c|c|c|}
\hline
\multicolumn{1}{|c|}{\textbf{Weather Condition}} & \multicolumn{2}{c|}{\textbf{Approaches}} \\ \cline{2-3} 
 & Augmented & Basic \\ \hline
Default & 0.26 ± 0.011 & \textbf{0.271 ± 0.011} \\
Fog & 0.186 ± 0.011 & \textbf{0.207 ± 0.013} \\
Night & 0.126 ± 0.007 & \textbf{0.136 ± 0.007} \\
Rain & \textbf{0.144 ± 0.008} & 0.14 ± 0.008 \\
Snow & 0.191 ± 0.007 & \textbf{0.197 ± 0.008} \\
\hline
\end{tabular}
\label{tab:results_yolo_size_n_map50}
\end{table}

\begin{table}[ht!]
\centering
\caption{Results of Faster RCNN evaluated with AP50 (mean ± std) in the Real-World Data Framework by weather conditions.}
\scriptsize
\setlength{\tabcolsep}{0.65mm}
\begin{tabular}{|c|c|c|c|c|c|}
\hline
\textbf{Weather} & \textbf{Approaches} & \multicolumn{4}{c|}{\textbf{Object Class}} \\ \cline{3-6} 
\textbf{Condition} &  & Walker & Vehicle & Traffic  & Traffic  \\
 &  &  &  &  Signs &  Lights \\ 
\hline
Default & Augmented & \textbf{0.669 ± 0.034} & \textbf{0.858 ± 0.010} & \textbf{0.687 ± 0.020} & \textbf{0.699 ± 0.027} \\
        & Basic     & 0.552 ± 0.043          & 0.798 ± 0.011          & 0.596 ± 0.022          & 0.595 ± 0.028          \\
\hline
Fog     & Augmented & \textbf{0.482 ± 0.061} & \textbf{0.776 ± 0.013} & \textbf{0.281 ± 0.024} & \textbf{0.258 ± 0.027} \\
        & Basic     & 0.477 ± 0.060          & 0.737 ± 0.014          & 0.270 ± 0.021          & 0.232 ± 0.026          \\
\hline
Night   & Augmented & \textbf{0.500 ± 0.033} & \textbf{0.608 ± 0.015} & \textbf{0.297 ± 0.018} & \textbf{0.247 ± 0.016} \\
        & Basic     & 0.415 ± 0.030          & 0.522 ± 0.016          & 0.266 ± 0.017          & 0.176 ± 0.015          \\
\hline
Rain    & Augmented & \textbf{0.371 ± 0.044} & \textbf{0.726 ± 0.017} & \textbf{0.219 ± 0.015} & 0.182 ± 0.022           \\
        & Basic     & 0.300 ± 0.040          & 0.704 ± 0.016          & 0.203 ± 0.015          & \textbf{0.185 ± 0.022} \\
\hline
Snow    & Augmented & \textbf{0.619 ± 0.028} & \textbf{0.713 ± 0.015} & \textbf{0.236 ± 0.014} & 0.173 ± 0.017           \\
        & Basic     & 0.572 ± 0.030          & 0.682 ± 0.015          & 0.218 ± 0.014          & \textbf{0.179 ± 0.018} \\
\hline
\end{tabular}
\label{tab:results_faster_rcnn_per_class_ap50_95_per_weather}
\end{table}

\begin{table}[ht!]
\centering
\caption{Results of YOLO Size M evaluated with AP50 (mean ± std) in the Real-World Data Framework by weather conditions.}
\scriptsize
\setlength{\tabcolsep}{0.65mm}
\begin{tabular}{|c|c|c|c|c|c|}
\hline
\textbf{Weather} & \textbf{Approaches} & \multicolumn{4}{c|}{\textbf{Object Class}} \\ \cline{3-6} 
\textbf{Condition} &  & Walker & Vehicle & Traffic  & Traffic  \\
 &  &  &  &  Signs &  Lights \\ 
\hline
Default & Augmented & 0.333 ± 0.036 & 0.576 ± 0.013 & \textbf{0.338 ± 0.022} & \textbf{0.240 ± 0.030} \\
        & Basic     & \textbf{0.359 ± 0.035} & \textbf{0.600 ± 0.013} & 0.314 ± 0.020 & 0.227 ± 0.029 \\
\hline
Fog     & Augmented & 0.236 ± 0.045 & \textbf{0.558 ± 0.016} & \textbf{0.140 ± 0.016} & 0.091 ± 0.019 \\
        & Basic     & \textbf{0.277 ± 0.049} & 0.515 ± 0.018 & 0.134 ± 0.017 & \textbf{0.095 ± 0.017} \\
\hline
Night   & Augmented & 0.173 ± 0.022 & 0.382 ± 0.018 & \textbf{0.100 ± 0.011} & \textbf{0.021 ± 0.006} \\
        & Basic     & \textbf{0.222 ± 0.023} & \textbf{0.392 ± 0.016} & 0.100 ± 0.011 & 0.015 ± 0.005 \\
\hline
Rain    & Augmented & 0.102 ± 0.028 & 0.498 ± 0.019 & \textbf{0.076 ± 0.008} & \textbf{0.054 ± 0.012} \\
        & Basic     & \textbf{0.151 ± 0.037} & \textbf{0.513 ± 0.018} & 0.057 ± 0.008 & 0.048 ± 0.011 \\
\hline
Snow    & Augmented & 0.212 ± 0.025 & 0.480 ± 0.017 & \textbf{0.106 ± 0.011} & 0.063 ± 0.011 \\
        & Basic     & \textbf{0.221 ± 0.027} & \textbf{0.518 ± 0.018} & 0.083 ± 0.009 & \textbf{0.072 ± 0.010} \\
\hline
\end{tabular}
\label{tab:results_yolo_size_m_per_class_ap50_95_per_weather}
\end{table}

\begin{table}[ht!]
\centering
\caption{Results of YOLO Size N evaluated with AP50 (mean ± std) in the Real-World Data Framework by weather conditions.}
\scriptsize
\setlength{\tabcolsep}{0.65mm}
\begin{tabular}{|c|c|c|c|c|c|c|}
\hline
\textbf{Weather} & \textbf{Approaches} & \multicolumn{4}{c|}{\textbf{Object Class}} \\ \cline{3-6} 
\textbf{Condition} &  & Walker & Vehicle & Traffic  & Traffic  \\
 &  &  &  &  Signs &  Lights \\ 
\hline
Default & Augmented & 0.199 ± 0.029 & 0.492 ± 0.013 & \textbf{0.217 ± 0.018} & 0.133 ± 0.023 \\
        & Basic     & \textbf{0.225 ± 0.027} & \textbf{0.506 ± 0.013} & 0.212 ± 0.019 & \textbf{0.141 ± 0.025} \\
\hline
Fog     & Augmented & 0.115 ± 0.035 & 0.452 ± 0.015 & 0.098 ± 0.014 & \textbf{0.080 ± 0.017} \\
        & Basic     & \textbf{0.191 ± 0.042} & \textbf{0.448 ± 0.015} & \textbf{0.121 ± 0.014} & 0.068 ± 0.017 \\
\hline
Night   & Augmented & 0.097 ± 0.017 & 0.317 ± 0.015 & \textbf{0.072 ± 0.009} & \textbf{0.019 ± 0.007} \\
        & Basic     & \textbf{0.139 ± 0.019} & \textbf{0.328 ± 0.015} & 0.066 ± 0.008 & 0.011 ± 0.004 \\
\hline
Rain    & Augmented & \textbf{0.082 ± 0.027} & 0.404 ± 0.017 & \textbf{0.049 ± 0.007} & 0.040 ± 0.010 \\
        & Basic     & 0.059 ± 0.021 & \textbf{0.418 ± 0.018} & 0.042 ± 0.006 & \textbf{0.041 ± 0.010} \\
\hline
Snow    & Augmented & 0.183 ± 0.020 & \textbf{0.438 ± 0.017} & \textbf{0.087 ± 0.009} & \textbf{0.057 ± 0.009} \\
        & Basic     & \textbf{0.219 ± 0.024} & 0.435 ± 0.016 & 0.080 ± 0.009 & 0.053 ± 0.009 \\
\hline
\end{tabular}
\label{tab:results_yolo_size_n_per_class_ap50_95_per_weather}
\end{table}

These lower results for the YOLO models also appear in the class-by-class analyses (Tables~\ref{tab:results_faster_rcnn_per_class_ap50_95_per_weather}--\ref{tab:results_yolo_size_n_per_class_ap50_95_per_weather}), where the difference between the two approaches is minimal. In contrast, the Faster R-CNN model demonstrates an increase in robustness under adverse conditions, achieving improvements such as a 0.07~mAP boost for pedestrian detection in night and rain settings compared to the \textit{Basic} approach. Thus, for models that achieve stable performance (such as Faster R-CNN), the increase in data improves robustness in the presence of adverse weather conditions.

\section{Conclusion}

This work investigated how tailored data augmentation with the Instruct Pix2Pix diffusion model can mitigate the performance gap of object detection systems operating in adverse weather conditions, which is of prominent importance in safety-relevant automotive systems due to the difficulties to collect datasets sufficiently complete, representative and balanced. In both the simulated (CARLA) and real-world (BDD100K + ACDC) domains, the proposed prompting strategy generated augmented images that plausibly resemble fog, night, rain, and snow scenarios. By systematically integrating these augmented images into the training process of object detectors, we observed improved robustness to weather-induced degradations.

In the simulated CARLA environment, augmentations reduced the accuracy drop when models were evaluated under challenging weather conditions. The improvements were consistent across Faster R-CNN and YOLO models, underscoring the effectiveness of Instruct Pix2Pix in producing realistic weather perturbations. For real-world data, Faster R-CNN benefited most from these augmentations, showing a smaller performance gap across different conditions compared to a baseline trained strictly on clear-weather images. While YOLO-based models did not display the same level of benefit in real-world evaluations, the results still suggest that carefully designed text prompts, combined with a filtering step to eliminate “hallucinated” samples, can help close the domain gap introduced by adverse weather.


Despite these improvements, challenges remain. First, the manual filtering of hallucinated images adds human oversight to the pipeline, leaving room for future automation through methods such as \cite{multimodalfilter}, which could ensure high-quality augmentations while minimizing manual intervention. Second, achieving a stronger benefit for single-stage YOLO models, especially under real-world conditions, may require more specialized training schemes. Lastly, expansion to additional weather phenomena, more diverse datasets, and refined hyperparameter tuning could further validate and strengthen the approach. Overall, this study demonstrates that diffusion-based image augmentations can be a practical and powerful tool to enhance the robustness of object detection in adverse weather conditions, a step towards safer and more reliable autonomous driving systems.

\section*{Acknowledgment}

This work has received funding from the European Union’s Chips Joint Undertaking (ChipsJU) framework, under grant agreement No 101139892 (EdgeAI-Trust), PCI2024-153417 financed by MICIU/AEI /10.13039/501100011033, and PLEC2023-010240 financed by MICIU/AEI /10.13039/501100011033 (CAPSUL-IA) and JDC2022-050313-I  funded by MCIN/AEI/10.13039/501100011033al by European Union NextGenerationEU/PRTR.

\end{document}